\documentclass[conference]{IEEEtran}
\IEEEoverridecommandlockouts
\usepackage{cite}
\usepackage{amsmath,amssymb,amsfonts}
\usepackage{booktabs,multirow,siunitx}
\usepackage{comment}
\usepackage{algorithm}
\usepackage{algpseudocode}
\usepackage{graphicx}
\usepackage{textcomp}
\usepackage{xcolor}
\usepackage{enumitem}
\usepackage{multirow}
\usepackage{booktabs}
\usepackage{siunitx}
\usepackage{placeins}
\usepackage{makecell}
\usepackage{threeparttable}
\usepackage{subcaption}
\usepackage[many]{tcolorbox}
\usepackage{import}
\usepackage{soul}
\def\BibTeX{{\rm B\kern-.05em{\sc i\kern-.025em b}\kern-.08em
    T\kern-.1667em\lower.7ex\hbox{E}\kern-.125emX}}

\makeatletter

\newcommand{\Rmnum}[1]{\expandafter@slowromancap\romannumeral #1@}

\makeatother

\usepackage{setspace}
\begin{document}

\title{TopoSizing: An LLM-aided Framework of Topology-based Understanding and Sizing for AMS Circuits\\
}

\author{
Ziming Wei\textsuperscript{1}, 
Zichen Kong\textsuperscript{2}, 
Yuan Wang\textsuperscript{2}, 
David Z. Pan\textsuperscript{3*}, 
Xiyuan Tang\textsuperscript{4*} \\
\textsuperscript{1}School of EECS, Peking University, Beijing, China \\
\textsuperscript{2}School of Integrated Circuits, Peking University, Beijing, China \\
\textsuperscript{3}Department of Electrical and Computer Engineering, The University of Texas at Austin, USA \\
\textsuperscript{4}Institute for Artificial Intelligence, Peking University, Beijing, China \\
*Corresponding authors: dpan@ece.utexas.edu, xtang@pku.edu.cn\\}

\maketitle
\begin{abstract}
Analog and mixed-signal (AMS) circuit design remains challenging due to the shortage of high-quality data and the difficulty of embedding domain knowledge into automated flows. Traditional black-box optimization achieves sampling efficiency but lacks circuit understanding, which often causes evaluations to be wasted in low-value regions of the design space. In contrast, learning-based methods embed structural knowledge but are case-specific and costly to retrain. Recent attempts with large language models (LLMs) show potential, yet they often rely on manual intervention, limiting generality and transparency.  
We propose \textbf{TopoSizing}, an end-to-end framework that performs robust circuit understanding directly from raw netlists and translates this knowledge into optimization gains. Our approach first applies graph algorithms to organize circuits into a hierarchical device–module–stage representation. LLM agents then execute an iterative hypothesis–verification–refinement loop with built-in consistency checks, producing explicit annotations. Verified insights are integrated into Bayesian optimization (BO) through LLM-guided initial sampling and stagnation-triggered trust-region updates, improving efficiency while preserving feasibility.  
On four real-world AMS circuits (OTA, FC-OTA, SACMP, LDO) in a 55-nm CMOS process, TopoSizing achieves \textbf{100\% correctness} in circuit understanding and parameter assignment. In constraint-satisfaction tasks, it delivers \textbf{1.4$\times$–4.8$\times$ higher sample efficiency} and \textbf{1.2$\times$–3.5$\times$ faster runtime} compared with strong baselines, while requiring 2–4$\times$ fewer LLM calls than prior LLM-based frameworks. Ablation studies confirm that reliable circuit understanding is essential for these gains.  
\end{abstract}


\section{Introduction}
Analog and mixed-signal (AMS) integrated circuits, such as operational amplifiers, comparators, and voltage references, remain indispensable in modern electronic systems. However, their design continues to rely heavily on expert intervention and domain-specific experience, making the process labor intensive and difficult to scale. Among the various stages, device sizing plays a particularly decisive role, as it determines key performance metrics. Yet, sizing is notoriously challenging: the design space is high-dimensional due to numerous interdependent parameters, and performance evaluation requires computationally expensive SPICE-level simulations. Together, these factors render sizing a critical but highly complex stage, one that this work directly aims to address.  

The most straightforward way to formulate sizing is as a black-box optimization problem, where device parameters are directly mapped to performance metrics. In this setting, heuristic algorithms\cite{SwarmSizing} and Bayesian Optimization (BO) and its variants~\cite{BOsizing,TRSizing,PVTSizing} have been widely applied, offering high sample efficiency and topology-agnostic applicability. However, black-box methods do not perform what expert designers routinely do, namely \textbf{circuit understanding}. Without the ability to interpret the structure and functionality of a circuit, these approaches cannot map topology into meaningful design constraints or guidance. As a result, they often fail to avoid wasting evaluations in unpromising regions that a human engineer would immediately exclude. This gap highlights the importance of circuit understanding as a necessary ingredient for practical automation.  

\textbf{Circuit understanding}, in this context, refers to the ability to reason from structure, parameters, and connectivity of circuits to infer their implications on performance and to translate this reasoning into design constraints or guidelines. The pursuit of such understanding has motivated researchers to explore learning-based methods. Early approaches trained surrogate models that mapped design parameters to performance metrics, effectively embedding domain knowledge into the optimization loop\cite{DNNOPT,APOSTLE,budak2021efficient,PPAAS}. Later efforts extended this idea with graph-based algorithms to explicitly encode circuit structure, capturing functional relationships between devices and modules\cite{GCN-RL,RGCNLDO,liu2021parasitic,DICE}. While these methods enriched the representation of circuits, they remain limited in practice: surrogate models are typically case-specific and require retraining whenever the topology changes, while dataset generation through costly simulations often outweighs the savings they provide. Consequently, despite their conceptual appeal, these approaches cannot scale or deliver efficient design automation.  

Up to this point, existing strategies reveal a fundamental dilemma: achieving topology generalizability, sampling efficiency, and embedded circuit understanding simultaneously has appeared nearly impossible. Naturally, the criteria for a form of \emph{circuit understanding} suitable for automated design can be summarized as follows: \textbf{(a) explicit preservation for reuse}, meaning the extracted knowledge is represented explicitly and stored in a form that can be reused across different design stages such as constraint formulation, testbench generation, or layout guidance; \textbf{(b) structural generality}, meaning the method can extend naturally to diverse circuit families and topologies without case-specific retraining; and \textbf{(c) transparency and verifiability}, meaning the intermediate roles of devices, modules, and stages are made explicit in a form that can be inspected, validated, and trusted.

However, achieving such circuit understanding is far from trivial: it requires a powerful model that can reliably interpret circuit topology, incorporate relevant domain knowledge, and correctly map this understanding into design requirements. Fortunately, the emergence of large-language models (LLMs) offers a promising path forward. Trained in vast and diverse corpora, LLMs possess broad prior knowledge that encompasses fundamental concepts in circuit theory and analog design. In addition, they exhibit a degree of reasoning capability, enabling them to interpret circuit connectivity, propose plausible sizing strategies, and support design-oriented inference. Moreover, their flexible input--output modalities make them easily extensible within existing workflows. Together, these characteristics suggest that LLMs could reconcile sample efficiency, topology transferability, and circuit knowledge integration within a single framework. This possibility has attracted increasing attention in the analog design automation community\cite{ADOLLM,LEDRO,analogcoder,Atelier}.

Motivated by these advantages, recent works have explored the use of LLMs in AMS electronic design automation (EDA). However, current approaches still suffer from significant limitations in scalability, generality, and transparency; a detailed review is provided in Section II. Together, the limitations of existing approaches can be traced to two main sources. The first lies in the unreliability of circuit understanding. LLMs are not naturally suited for reasoning over graph-structured data\cite{gpt4graph} such as circuits, and raw netlists pose particular difficulties. Unlike designer-authored netlists, which often contain implicit hints such as ordered indexing, hierarchical organization, or meaningful node names, raw netlists, especially those extracted automatically from images or heterogeneous sources, lack such structure. Their nodes typically do not have informative labels, their ordering is arbitrary, and functional modules are not delineated. Therefore, it is unrealistic to assume that an LLM can directly parse these netlists into the correct functional roles\cite{Comprehend-benchmark}. Achieving reliable understanding requires working strictly from topology-only information and deriving a robust, high-quality interpretation akin to how human engineers decompose circuits.  
The second difficulty lies in how to effectively leverage the attained circuit understanding to achieve tangible improvements in optimization efficiency. This requires first the choice of a principled backbone optimization algorithm and then the design of appropriate intervention mechanisms. However, existing LLM-based works lack consensus on what constitutes the most effective integration strategy, and the approaches adopted so far remain relatively coarse. For example, some studies bypass circuit knowledge entirely by invoking an optimization algorithm directly, while others let the LLM propose candidate sampling points or prune the search space based on the current optimization state. A more systematic methodology is needed to translate circuit understanding into design efficiency, along with comparative studies to evaluate different intervention strategies.

In response to the above challenges, we propose an end-to-end framework that spans the analog front-end design flow, taking a raw netlist as input and producing a sized circuit as output, with circuit understanding accomplished in the process and preserved as intermediate knowledge in textual or annotated form. The framework begins with raw netlists and performs robust circuit understanding based solely on topology, assisted by graph algorithms and LLMs. This circuit understanding is not only essential for guiding subsequent design tasks within our framework but also tangible in the sense that the extracted knowledge can be retained for potential reuse in other applications. On top of this, sizing serves both as a strengthened optimization stage and as a validation mechanism: the acceleration achieved in sizing demonstrates the practical benefits of circuit understanding, while the correctness of sizing outcomes in turn substantiates the reliability of the proposed circuit understanding process.  

Our main contributions are summarized as follows:  
\begin{itemize}
    \item We present a unified end-to-end framework for analog front-end design that integrates LLMs into the workflow. Starting from raw netlists, the framework performs topology-based circuit understanding and leverages the extracted annotations to enhance downstream tasks, with the knowledge preserved for potential reuse.  
    \item We propose a graph-assisted methodology for reliable circuit understanding. Circuit information is systematically extracted into a hierarchical representation across three levels--component, module, and stage--and the LLM executes iterative reasoning in a hypothesis-verification-refinement loop, improving robustness while ensuring interpretability.  
    \item We introduce an efficient integration of LLM-based circuit understanding with BO. By incorporating LLM-guided initial sampling and stagnation-triggered trust-region updates, our approach improves BO efficiency through simple yet effective mechanisms.  
    \item We validate the framework on four real-world circuit cases, where TopoSizing achieves \textbf{100\% correctness} in both circuit understanding and parameter assignment. In constraint satisfaction tasks, our method delivers \textbf{1.4$\times$ to 4.8$\times$ higher sample efficiency} and \textbf{1.2$\times$ to 3.5$\times$ faster runtime} compared with traditional optimization baselines.  
\end{itemize}

\section{Preliminaries}

\subsection{AMS Circuit Understanding and Sizing}
From the above discussion, we note that our entire workflow begins with establishing a reliable \emph{circuit understanding}. To evaluate this crucial step, we introduce two indicators: (i) the \emph{classification accuracy} of devices and modules, which reflects whether circuit components are assigned to the correct functional roles; and (ii) the \emph{correctness of design-constraint or parameter assignment}, which measures whether the extracted circuit information can be faithfully translated into sizing-related guidance.

To further demonstrate the practical value of circuit understanding, we take AMS circuit sizing as a downstream task.  
AMS sizing refers to the process of tuning device-level parameters—such as transistor widths, lengths, or bias currents—so that the resulting circuit meets all performance specifications such as gain, bandwidth, power consumption, and stability.  
Successful sizing requires searching over a high-dimensional design space, and its efficiency can be significantly improved when guided by accurate circuit understanding.  
Thus, sizing serves both as a key design objective and as an auxiliary validation of the usefulness of our framework.

\subsection{Problem Formulation for Sizing}
AMS circuit sizing can be cast as a constrained optimization problem.  
For the feasibility scenario, the task is to identify a design $\mathbf{x}\in\mathcal{X}$ that meets all performance specifications:
\begin{equation}
\text{find } \mathbf{x} \in \mathcal{X} 
\quad \text{s.t. } F_i(\mathbf{x}) \ge C_i, \;\; \forall i \in \{1,\dots,N_c\},
\label{eq:feasibility}
\end{equation}
where $F_i(\mathbf{x})$ is the $i$-th performance metric, $C_i$ its specification, and $N_c$ the number of constraints.  

For the single-objective constrained optimization, one target metric $F_t(\mathbf{x})$ is maximized while ensuring feasibility of all other constraints:
\begin{equation}
\max_{\mathbf{x}\in\mathcal{X}} \;\; F_t(\mathbf{x})
\quad \text{s.t. } F_i(\mathbf{x}) \ge C_i, \;\; \forall i \ne t.
\label{eq:singleobj}
\end{equation}
This formulation highlights the two experimental settings considered in our framework: 
(i) pure feasibility search for rapid constraint satisfaction, and 
(ii) performance-driven optimization under design constraints.

\subsection{LLMs for AMS EDA}
\label{llm_related}
Existing approaches of utilizing LLMs in AMS EDA can be broadly divided into two categories.

The first line of work develops domain-specific LLMs tailored for AMS design tasks. These methods aim to incorporate circuit-specific knowledge into pretraining or fine-tuning, but they are constrained by the scarcity of large, high-quality datasets. Despite auxiliary efforts such as extracting netlists from schematics or images~\cite{AMSnet,Image2net}, automatically generating sizing scripts~\cite{AnaSizeCoder}, and constructing test benches~\cite{AutoTB}, ensuring correctness in circuit interpretation and parameter allocation still requires substantial manual validation. This bottleneck limits both scalability and reliability.

The second line of work attempts to directly apply general-purpose LLMs within the design process. In the context of sizing, for example, retrieval-augmented generation (RAG) and multi-agent discussion have been explored to provide design guidance~\cite{Atelier,ampagent}. Other studies let LLMs intervene in optimization directly~\cite{ADOLLM,LEDRO}. However, these approaches typically process raw netlists without additional structure, which restricts applicability to simple topologies. Moreover, their interpretations are rarely verified, leaving the process opaque and lacking transparency. Even in optimization integration, strategies remain coarse and inconsistent: some pipelines bypass model intervention entirely~\cite{Atelier}, while others only allow the LLM to propose a few candidate samples per iteration~\cite{ADOLLM}.  

Nevertheless, these efforts highlight the larger potential of LLM-in-the-loop workflows: as commercial models continue to improve, such pipelines could, in principle, achieve even stronger design performance. The key challenge, however, lies in ensuring interpretability and reliability. In particular, existing approaches rarely verify whether the LLM’s circuit understanding is correct or whether its design guidance is appropriate—both of which are crucial for trustworthy automation. This gap motivates our work, where we propose a framework that not only leverages LLMs for circuit understanding but also incorporates mechanisms to validate and refine their interpretations.

In summary, while these works demonstrate the potential of LLMs in AMS EDA, they also reveal clear limitations in scalability, generality, and verifiability. Addressing these gaps motivates the systematic framework proposed in this paper.

\subsection{Circuit Representation: Requirements and Choices}

A central question in enabling LLMs for AMS EDA is how to represent circuits in a form that is both \textbf{information-preserving} and \textbf{LLM-compatible}. On the one hand, the representation must preserve all relevant structural and parametric information, since any loss of connectivity or device attributes would compromise the fidelity of subsequent analysis. On the other hand, the representation must be interpretable by LLMs, whose pretraining distribution is dominated by natural language rather than specialized graph data or domain-specific syntaxes.

From the perspective of information completeness, electrical circuits are naturally graphs: devices and terminals can be modeled as nodes, while electrical connections form edges. Such graph-based representations are widely used in learning-based electronic design automation (EDA), particularly with graph neural networks (GNNs)~\cite{DICE,GCN-RL,RGCNLDO}. They provide a lossless encoding of circuit topology and attributes. However, these representations are not well aligned with LLMs. Raw adjacency matrices or edge lists are high-dimensional, order-sensitive, and difficult to tokenize effectively; linearizing them introduces artificial ordering biases; and functional relationships such as differential pairs or current mirrors remain implicit~\cite{gpt4graph}. Consequently, although graph representations excel at preserving information, they do not naturally support LLM reasoning.

Another direction is to augment the raw netlist with \textbf{semantic annotations}. For instance, designers may explicitly mark functional groups. Such annotations improve interpretability but typically require substantial manual effort, making them difficult to scale across large design spaces.  
A related approach is to employ code-like domain-specific languages (DSLs) for circuit description. These languages introduce specialized syntax to represent modules and hierarchical structure in a more formal and programmable way, akin to hardware description languages\cite{TED}. While such DSLs can provide concise and structured descriptions, they rely on conventions that pretrained LLMs are not exposed to, and thus may not be easily understood without additional fine-tuning or task-specific adaptation.  
In summary, both directions have demonstrated effectiveness within their respective application scenarios, but neither fully satisfies the dual requirement of being information-encompassing and LLM-compatible. 
This gap motivates our approach, which aims to balance these trade-offs by transforming graph-extracted circuit information into structured natural-language descriptions that preserve essential topology while remaining aligned with LLMs’ strengths.

\section{Methodology}
\begin{figure*}[t!]
  \raggedright 
  \includegraphics[width=0.98\textwidth]{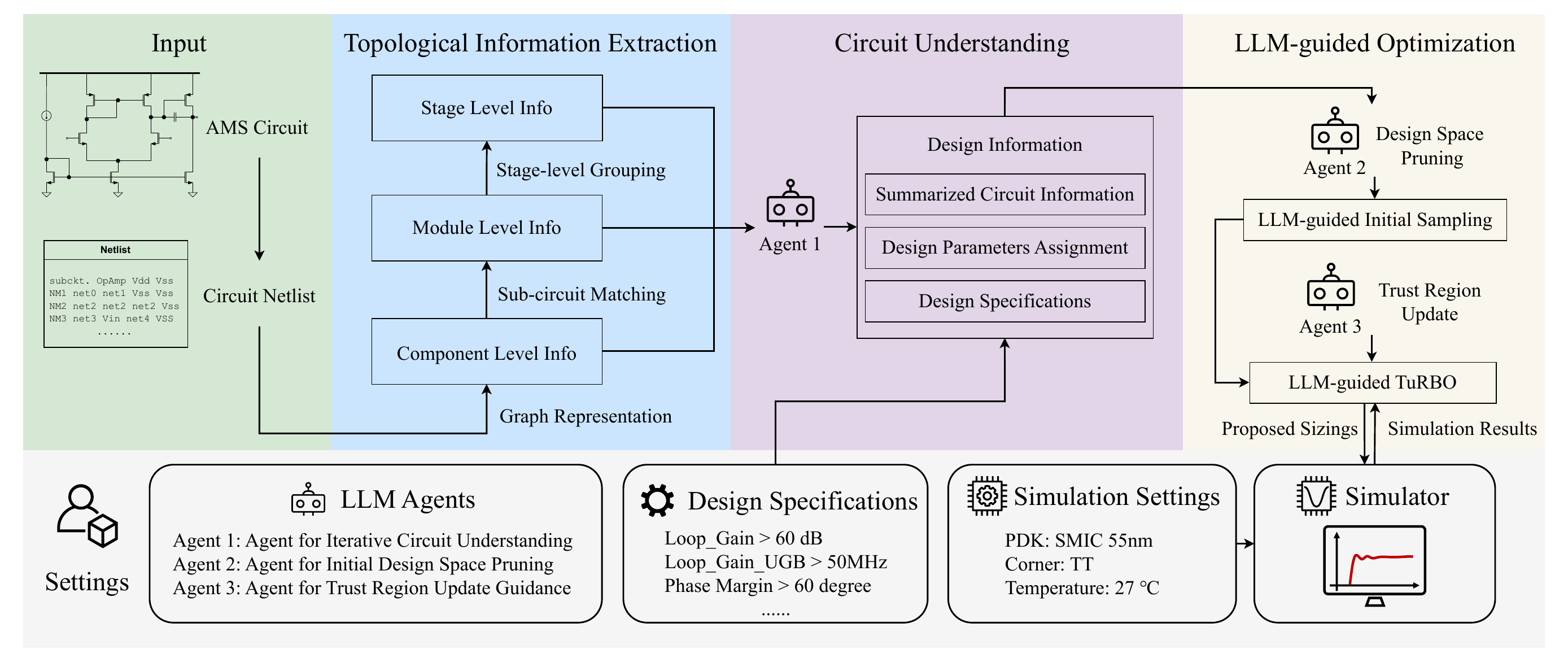}
  \caption{The work flow of TopoSizing}
  \label{fig:framework}
\end{figure*}
\subsection{Overview}

In this section, we describe the detailed implementation of \textbf{TopoSizing}, a fully automated framework that integrates graph-based processing with LLMs to achieve reliable circuit understanding and sample-efficient optimization (Fig.~\ref{fig:framework}). The workflow is carried out in three stages. First, raw netlists are transformed into hierarchical graph representations that capture component, module, and stage-level structure. Based on this representation, LLM agents iteratively perform circuit understanding, producing explicit and verifiable annotations of functional roles. Finally, this circuit understanding is coupled with an adapted BO, where LLM guidance refines the design space and gives instructions on trust region update. By moving from structural encoding to functional interpretation and ultimately to optimization, TopoSizing integrates all stages into a coherent end-to-end pipeline.

\subsection{Topological Information Extraction}

The first step toward reliable circuit understanding is to choose
an LLM- and algorithm-friendly circuit representation that is both expressive and easy to process.
Therefore, we attempt to construct a structured container that captures the circuit at three hierarchical levels: \emph{component}, \emph{module}, and \emph{stage}. 
This organization preserves both fine-grained details and higher-level functional context, making the representation both expressive and easy to query. 
An additional advantage is that hierarchical abstraction effectively reduces the connection complexity inherent in graph-structured netlists. Both advantages of this structured graph container are particularly beneficial in improving the robustness of LLM-based circuit understanding\cite{language_is_all_a_grah_needs}. 
The design of this hierarchy is inspired by how analog designers analyze circuits: starting from the supply to divide the circuit into stages, 
then identifying common module sub-circuits to quickly infer functionality, and finally inspecting individual component connections to verify details.

\subsubsection{\textbf{Circuit to graph transformation}}
Circuits naturally exhibit the properties of a graph: devices and nets form nodes, and their connections define edges. Leveraging this intrinsic structure, we transform a raw netlist into a structured, semantically enriched graph in a fully automated way. This process is almost “free” since it reuses the connectivity already encoded in the netlist and produces a unique, lossless representation that preserves all original information.

\[
\begin{aligned}
\mathcal{G} &= (\mathcal{V}, \mathcal{E}), \\
\mathcal{V} &= \mathcal{V}_{\text{dev}} \cup \mathcal{V}_{\text{net}}, \\
\mathcal{V}_{\text{dev}} &= \{\, v_i \mid \tau(v_i) \in \{\text{NMOS}, \text{PMOS}, R, C, I, V\} \,\}, \\
\mathcal{V}_{\text{net}} &= \{\, v_j \mid \tau(v_j) \in \{\text{net}, \text{GND}, \text{VDD}\} \,\}, \\
\mathcal{E} &\subseteq \{\, (v^{\text{dev}}_i, v^{\text{net}}_j, \ell) \mid \ell \in \mathcal{L} \,\}, \\
\mathcal{L} &= \{ G, S, D, R, C, I, V \}.
\end{aligned}
\]

Here, $\mathcal{G}$ is the circuit graph with node set $\mathcal{V}$ and edge set $\mathcal{E}$. 
The node set is partitioned into device nodes $\mathcal{V}_{\text{dev}}$ and net nodes $\mathcal{V}_{\text{net}}$, 
where $\tau(v_i)$ denotes the node type. 
Each edge $\varepsilon \in \mathcal{E}$ connects a device node to a net node with label $l \in \mathcal{L}$, 
where $l$ indicates terminal roles or device types.

Edges $\varepsilon\in\mathcal{E}$ connect device nodes to net nodes with a label \(\ell \in \mathcal{L}\). For MOSFETs, labels indicate terminal roles: Gate (G), Source (S), Drain (D). For other devices, the label corresponds to the device type. Since connections only occur between devices and nets, the resulting graph is bipartite, capturing both topology and port-level semantics.

This structured encoding offers two key benefits: (1) it is lossless, preserving every device, connection, and functional role, and (2) it is directly compatible with graph-based algorithms. As a result, it serves as an ideal foundation for subsequent topology analysis, module recognition, and optimization.

\subsubsection{\textbf{Sub-circuit Matching}}

\begin{figure*}[t]
  \raggedright 
  \includegraphics[width=0.98\textwidth]{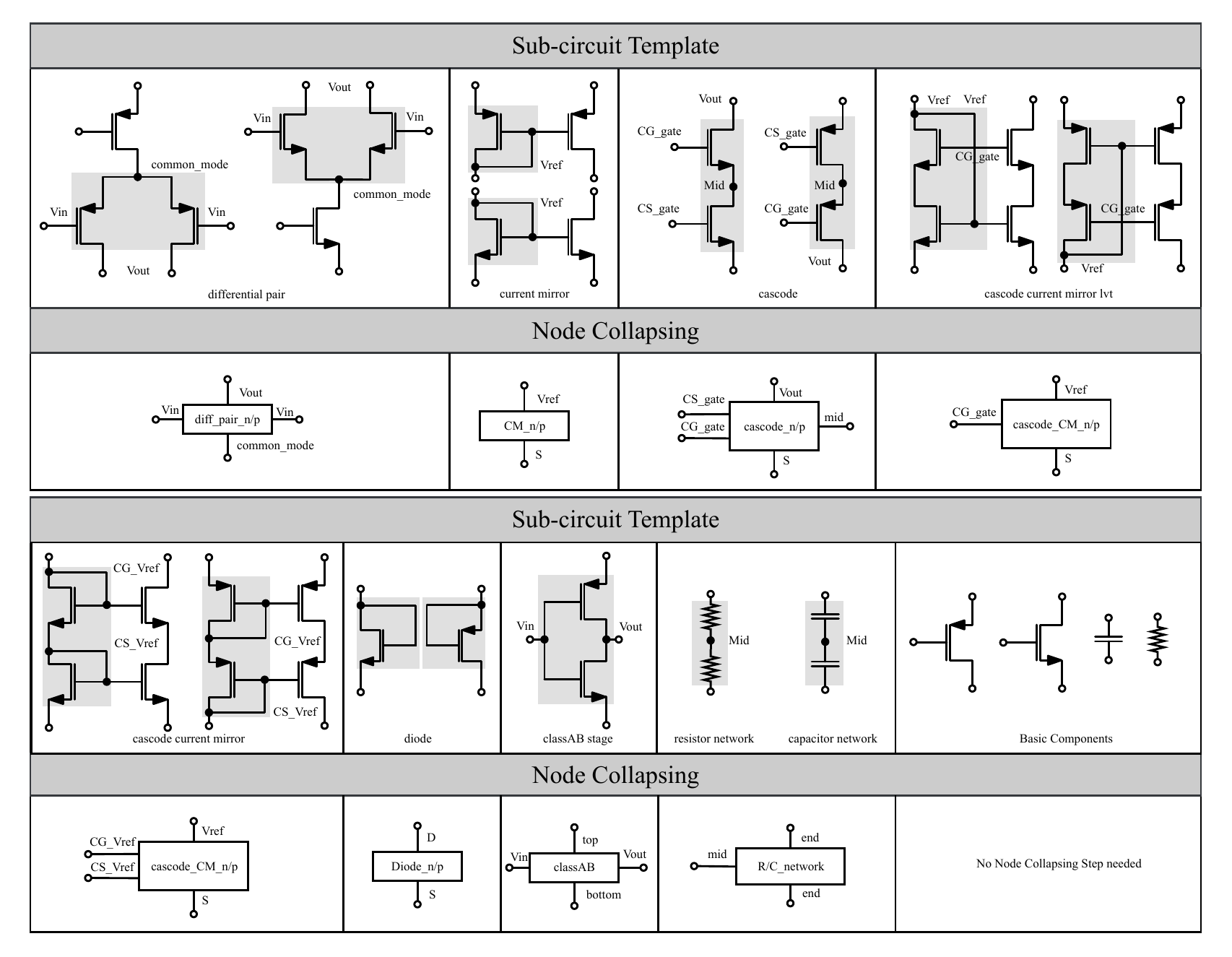}
  \caption{The sub-circuit templates and supernode collapsing. The connection can be depicted accurately in text by embedding names into the edges and nodes.}
  \label{fig:subckt}
\end{figure*}

To further simplify the circuit graph while retaining meaningful functional structure, we take advantage of a characteristic unique to analog circuits: the frequent reuse of well-established basic modules such as differential pairs, current mirrors, and cascode stages. This is where our graph-based representation shows its strength: once a circuit is represented as a graph, we can directly apply subgraph algorithms to identify and manipulate these recurring patterns.

In our approach, we first construct a library of commonly used analog building blocks. Each building block is a labeled subgraph in which both node types and edge types are explicitly specified, encoding canonical topological patterns. These templates not only guide the matching process, but also link specific transistors and devices to their functional roles within a module, enabling us to later reason about how design parameters influence performance. The initial library covers differential pairs, current mirrors, cascode stages, cascode current mirrors, Class-AB stages, and diode-connected MOS devices. These modules are chosen because they constitute the fundamental building blocks of analog design, are repeatedly reused across circuits, and together span almost all practical topologies\cite{razavi2005design,OASYS}. The library can also be readily extended to more complex structures.

Based on the graph representation of the circuit, we perform subgraph isomorphism to detect template matches. When a match is found, we replace the matched devices with a single supernode representing the identified module. Net nodes, which define the electrical interconnection points, are preserved to maintain global connectivity. This replacement does more than reduce complexity, it also embeds new semantic information into the graph. By renaming the supernode and adjusting the edge types at its boundary, we encode both the module’s identity and its port-level roles directly into the graph’s structure. This makes later queries trivial—design tools or LLM agents can retrieve functional details simply by inspecting node and edge labels.

\subsubsection{\textbf{Stage-level Grouping}}

\begin{figure*}[t]
  \raggedright 
  \includegraphics[width=1\textwidth]{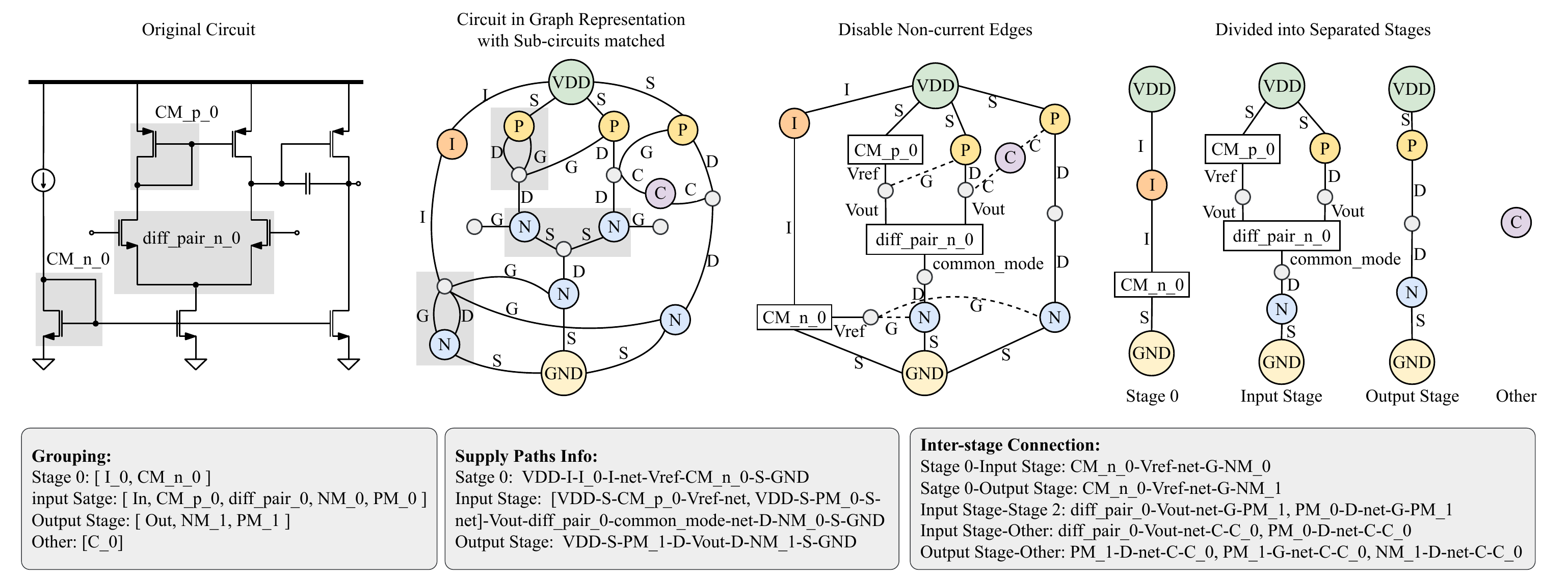}
  \caption{Example of the grouping process of the matched circuit. The circuit topology can be then presented in text as shownn in the lower half of the figure.}
  \label{fig:grouping}
\end{figure*}

Even after subcircuit matching and module abstraction, the circuit graph can remain highly complex: it may still contain a large number of nodes, no clear hierarchical organization, and many connections that a designer would not consider functionally important. To further organize this structure, we take inspiration from a common principle in analog circuit design, dividing the circuit into stages, and formulate a rule-based method for grouping components based on whether a continuous current can flow through them. 

The core idea is that components sharing a continuous current supply path typically form a unified functional stage, operating under the same biasing conditions and serving a specific electrical role in the overall system. To implement this, we first reduce the circuit graph by removing edges that do not conduct current, such as those corresponding to capacitors or MOSFET gate connections. This results in a pruned subgraph:
\[
\mathcal{G}_{\text{i}} = (\mathcal{V}_{\text{i}}, \mathcal{E}_{\text{i}}) \subseteq \mathcal{G}.
\]

Within this current-conduction graph, a \emph{conduction path} is defined as a sequence of alternating device and net nodes that connect the supply node \( v_{\text{VDD}} \in \mathcal{V}_{\text{net}} \) to the ground node \( v_{\text{GND}} \in \mathcal{V}_{\text{net}} \):
\[
\mathcal{P}_k = (v_1, e_1, v_2, \dots, v_n), \quad \text{with } v_1 = v_{\text{VDD}},\ v_n = v_{\text{GND}}.
\]
All such paths are enumerated and collected:
\[
\mathcal{P} = \{ \mathcal{P}_1, \mathcal{P}_2, \dots, \mathcal{P}_K \}.
\]

Rather than treating each conduction path independently, we merge any two paths that share an internal node:
\[
\mathcal{P}_i \sim \mathcal{P}_j \iff \exists v \in \mathcal{P}_i \cap \mathcal{P}_j, \quad v \notin \{v_{\text{VDD}}, v_{\text{GND}}\}.
\]
This defines an equivalence relation over \(\mathcal{P}\), and the union of equivalent paths forms a single stage:
\[
\mathcal{S}_l = \bigcup_{\mathcal{P}_k \in [\mathcal{P}]_l} \mathcal{P}_k, \quad l = 1, \dots, L.
\]

Once the stages \(\{\mathcal{S}_1, \dots, \mathcal{S}_L\}\) are identified, we define a stage-level interconnection graph:
\[
\mathcal{G}_{\text{stage}} = (\mathcal{S}, \mathcal{E}_{\text{inter}}),
\]
where stages are divided and edges capture valid inter-stage signal connections. Formally:
\[
(\mathcal{S}_i, \mathcal{S}_j) \in \mathcal{E}_\text{inter}
\quad\textbf{iff}\quad
\begin{cases}
\exists\, v \in \mathcal{V}_\text{net}, \\[2pt]
\exists\, u_i \in \mathcal{S}_i \cap \mathcal{V}_\text{dev}, \\[2pt]
\exists\, u_j \in \mathcal{S}_j \cap \mathcal{V}_\text{dev}, \\[4pt]
(u_i, v) \in \mathcal{E} \ \land\ (u_j, v) \in \mathcal{E}.
\end{cases}
\]

This ensures that stages are connected only through shared nets, with no device-level overlap, aligning with the modular and hierarchical nature of analog circuit design. We also annotate stages containing primary input or output terminals, enabling the generation of a stage-level signal flow model.  This hierarchy serves as a bridge between low-level device representation and higher-level functional architecture, providing a clear, query-friendly structure for subsequent analysis and LLM interaction.

The outcome of this step is a hierarchical graph-based container for circuit information. At the lowest level, it records individual transistors and their precise electrical connections; at the next level, these devices are grouped into recognized sub-circuits; and at higher levels, sub-circuits are organized into functional stages. This layered organization preserves complete connectivity while making it easy to query relationships—whether between specific ports, within a module, or across stages. In this way, the graph serves as an accessible and structured repository from which later stages of the framework can retrieve exactly the information they need, without losing any of the detail present in the original netlist.

\subsection{\textbf{Circuit Understanding}}
\begin{figure}[]
  \includegraphics[width=0.49\textwidth]{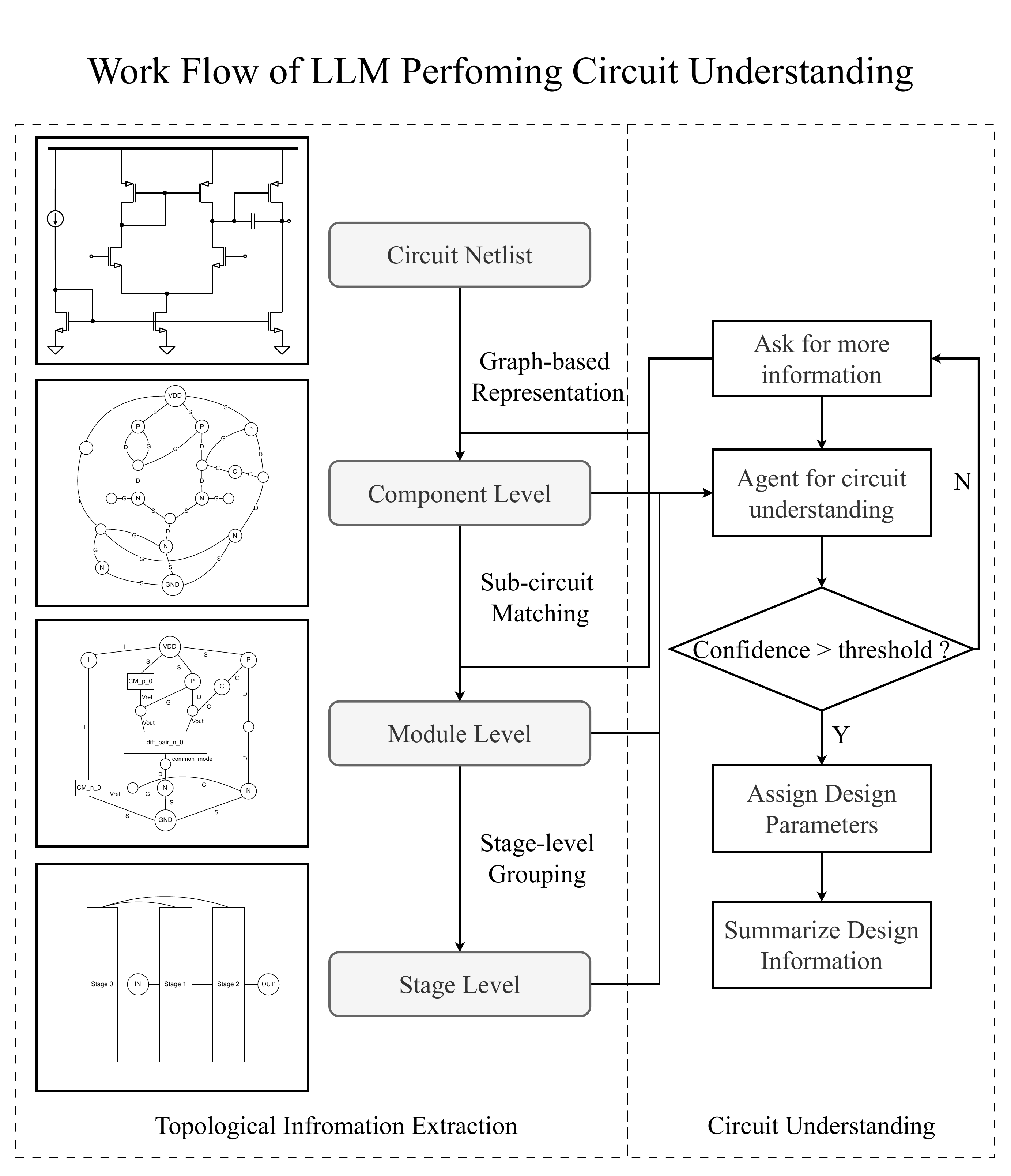}
  \caption{LLM agent performs circuit understanding based on the hierarchical topology information extracted and stored in the previous steps.}
  \label{fig:topo-understanding}
\end{figure}
\subsubsection{Iterative Circuit Understanding via LLM Agents}
To achieve an accurate and robust understanding of the circuit’s functional behavior, our framework employs dedicated LLM agents that iteratively analyze and refine circuit interpretations. Initially, an LLM agent processes structured circuit representations, including graph-based topology and hierarchical module information, to generate a functional analysis.

Crucially, the agent incorporates an internal confidence assessment mechanism, whereby it evaluates the certainty of its interpretations for each node, connection, or functional relationship. To ensure reliability, our model further applies a simple checklist\cite{TICK} during recognition, ensuring that all device functions are successfully identified, that the functional roles hypothesized for devices within each stage are consistent with their interconnections, and that the overall circuit hierarchy and connectivity align with the inferred functional assignments. Only after these checks are satisfied does the agent output its result as confident. When the agent detects ambiguities or low-confidence regions, it generates targeted queries to disambiguate these uncertain elements. These queries are answered through subsequent interactions, enabling the model to iteratively refine its understanding.

The iterative process continues until the agent’s confidence surpasses a predefined threshold, at which point the analysis is considered sufficiently accurate and the iteration terminates. This confidence-driven stopping criterion ensures that the model dedicates more effort to complex or uncertain circuit regions, while avoiding unnecessary computations on well-understood portions.

Throughout this workflow, a coordination agent collects and organizes the progressively refined insights into structured prompts. These comprehensive representations encode the hierarchical and functional properties of the circuit, facilitating downstream tasks such as optimization guidance and design space exploration. By integrating confidence-aware iterative analysis, the system achieves a high-fidelity understanding of complex circuit structures in an efficient and scalable manner.

\subsubsection{LLM-Assisted Design Parameter Assignment}

In addition to functional understanding, the framework leverages LLM agents to assign design parameters in a symmetry-aware manner. Many analog circuits contain inherently symmetric or replicated structures, such as differential pairs, current mirrors, or cascaded bias branches, where corresponding devices share identical or proportionally related sizing parameters. By automatically detecting these symmetric or functionally equivalent components from the hierarchical circuit representation, the LLM enforces consistent parameter assignments across them. This process significantly reduces the effective dimensionality of the design space, as multiple variables can be tied together under symmetry constraints rather than optimized independently. Such parameter tying mirrors common practices in manual sizing, where designers pre-define equality or ratio relationships to maintain matching characteristics and simplify the optimization process. Incorporating this capability into the automated framework not only accelerates convergence but also ensures that the generated designs adhere to well-established analog design principles.

\subsection{LLM-guided Optimization}

Our ultimate goal is to leverage the circuit knowledge learned by the LLM to accelerate device sizing. BO is a classical and highly efficient search algorithm in device sizing, making it a natural choice as our optimization backbone. To inject domain knowledge, we leverage the LLM’s circuit reasoning to intervene at two critical factors influencing BO performance: enhancing initial sampling quality through conservative space pruning, and refining trust region updates via selective guidance when stagnation occurs. This targeted integration preserves BO’s efficiency while steering the search with domain-informed direction.

\subsubsection{LLM-Guided Initial Sampling}

The proposed LLM-guided optimization framework addresses the device sizing problem in analog circuit design, where optimal sizing parameters are strongly dependent on the specific technology library. Due to the inherent variations in device models, sizing solutions cannot be directly transferred across technologies. Consequently, the LLM cannot generate high-performance design points without prior knowledge; instead, it must rely on representative samples to infer promising regions of the design space.

To this end, the optimization process begins with an initial sampling stage. Based on the structural and functional information extracted in the previous stages, the LLM performs a loosely constrained search space pruning. The pruning is intentionally conservative to avoid excluding potentially viable designs. For example, only basic feasibility constraints are imposed, such as ensuring that the width-to-length ratio (\(W/L\)) of differential pairs exceeds~5. This eliminates parameter configurations that are clearly infeasible while retaining a broad search range.

The initial sampling is conducted primarily within the LLM-pruned subspace, ensuring that most samples are concentrated in regions more likely to yield good performance. However, to preserve global exploration capability, a controlled fraction of samples is drawn from outside the pruned region. This hybrid sampling strategy provides the optimizer with an informed and more effective initial dataset, while avoiding premature over-restriction of the search domain.

\begin{algorithm}[!htbp]
\caption{Initial Sampling}\label{alg:alg2}
\begin{algorithmic}[1]
\State \textbf{Input:}
\Statex $D,D_{P}$ - original and pruned design space
\Statex $\alpha$ - Pruned space sampling ratio

\State Sample $\{x_i^p\} \sim \mathcal{LHS}(P)$ 
\State Sample $\{x_i^r\} \sim \mathcal{LHS}(D \setminus P)$ 
\State $x_i \gets \{x_i^p\} \cup \{x_i^r\}$ with ratio $\alpha:(1-\alpha)$
\State Evaluate $FoM_i \gets \{FoM(x_i)\}$
\State \textbf{return} $ (x_i,FoM_i) $
\end{algorithmic}
\end{algorithm}

\subsubsection{Stagnation Triggered Trust Region Update}

Following the initial sampling, the optimization proceeds using the TuRBO (Trust Region BO) algorithm\cite{Turbo}, which adaptively explores the high-dimensional design space in parallel. While TuRBO efficiently exploits local structure, it may become trapped in sub-regions when performance improvements plateau. To address this, the LLM is incorporated as an adaptive guide for trust region adjustment.

The LLM is not invoked at every iteration to reduce computational and resource overhead. Instead, its intervention is triggered by a \textit{stagnation criterion}, specifically, when the best observed performance fails to improve over a predefined number of consecutive iterations. Upon activation, the LLM analyzes the accumulated performance data together with the topological and functional circuit information. It then generates refined constraints or directional guidance to adjust the trust region boundaries, steering the optimizer toward unexplored yet promising regions of the parameter space.

This selective intervention strategy strikes a balance between the exploitation capacity of TuRBO and the global redirection enabled by the LLM. By limiting LLM queries to moments of optimization stagnation, the framework minimizes unnecessary model calls while maintaining the ability to dynamically reorient the search toward high-performance regions.

\begin{algorithm}[!t]
\caption{LLM-Guided TuRBO }\label{alg:llm_turbo}
\begin{algorithmic}[1]
\State \textbf{Input:} $N_{\text{init}},N_{\text{iter}},K,\ \alpha_{\text{inc}}>1,\ \alpha_{\text{dec}}\in(0,1),\ r_{\min}<r_{\max}$
\State Randomly sample $N_{\text{init}}$ points, fit $\mathcal{GP}_0$; set $x^\star=\arg\max \mathrm{FoM}$, $\mathrm{FoM}^\star=\max \mathrm{FoM}$, $r_0\in[r_{\min},r_{\max}]$, \textit{no\_imp}$=0$
\For{$t=0$ to $N_{\text{iter}}-1$}
  \State Within $\mathcal{TR}_t=\{x:\|x-x^\star\|_\infty\le r_t\}$, simulate and update $\mathcal{GP}_{t+1}$
  \State $y_{\max}=\max \mathrm{FoM}(X_t)$, $x_{\text{new}}=\arg\max \mathrm{FoM}(X_t)$
  \State \textbf{If $y_{\max}>\mathrm{FoM}^\star$:} $(x^\star,\mathrm{FoM}^\star)=(x_{\text{new}},y_{\max})$,  $r_{t+1}=\min(\alpha_{\text{inc}}r_t,r_{\max})$, \textit{no\_imp}=0; 
    \State else: $r_{t+1}=\max(\alpha_{\text{dec}}r_t,r_{\min})$, \textit{no\_imp}+=1
  \State If \textit{no\_imp}$\ge K$: query $\mathcal{LLM}$, receive $(\tilde x,\tilde r)$, set $x^\star\leftarrow\tilde x$, $r_{t+1}\leftarrow\operatorname{clip}(\tilde r;r_{\min},r_{\max})$, \textit{no\_imp}=0
\EndFor
\State \textbf{Return:} $(x^\star,\mathrm{FoM}^\star)$
\end{algorithmic}
\end{algorithm}

\begin{figure*}[t!]
  \raggedright 
  \includegraphics[width=1\textwidth]{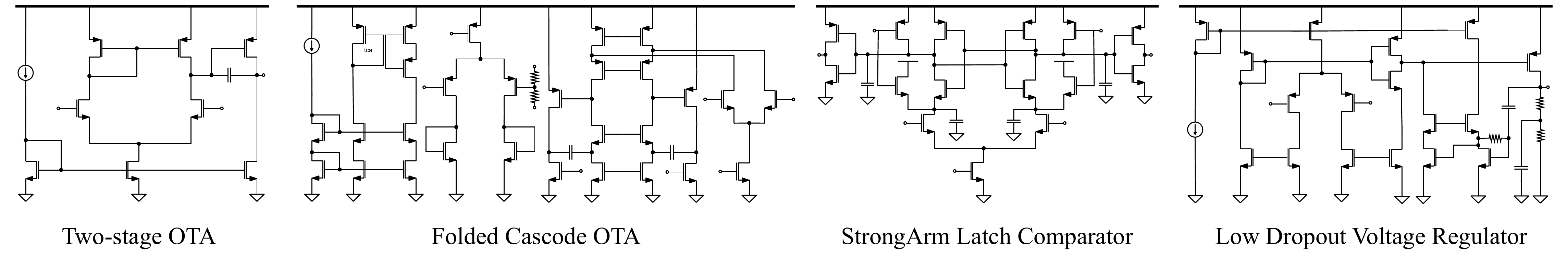}
  \caption{Schematics of four real-world circuits.}
  \label{fig:testcase}
\end{figure*}
\section{Experimental Results}
\subsection{Testcases}
We select four real-world analog circuits to evaluate the design efficiency of TopoSizing: a two-stage operational transconductance amplifier (OTA), a folded-cascode operational transconductance amplifier (FCOTA), a StrongArm latch comparator (SACMP), and a low-dropout regulator (LDO). All circuits are implemented using the commercial SMIC 55nm CMOS process and are derived from practical engineering applications, ensuring their real-world relevance.

As illustrated schematically in Fig.~\ref{fig:testcase}, these circuits are carefully chosen to cover a wide range of topologies, performance specifications, and design space complexities. The following sections detail the rationale behind the selection of these benchmarks and describe the corresponding experimental setups.

\textbf{OTA:} This circuit includes 11 design parameters: 4 transistor widths, 4 transistor lengths, 2 transistor ratios, and 1 capacitance value. The target specifications are:
\begin{equation}
C=
\begin{cases}
    \text{Gain}\geq40\text{ dB,}&\text{GBW}\geq50\text{ MHz}\\
    \text{Phase margin}\geq60^{\circ}\text{,}&\text{Power}\leq0.5\text{ mW}
\end{cases}
\end{equation}
\textbf{FCOTA:}  This circuit contains 20 design parameters: 7 transistor widths, 7 lengths, 4 ratios, and 2 capacitance values. The circuit comprises 33 components, including both core stages and biasing networks. The target specifications are :
\begin{equation}
C=
\begin{cases}
    \text{Gain}\geq100\text{ dB,}&\text{GBW}\geq30\text{ MHz}\\
    \text{Phase margin}\geq60^\circ\text{,}&\text{Power}\leq1\text{ mW}\\
     \text{PSRR}\geq100\text{ dB,}&\text{CMRR}\geq100\text{ dB}\\
     \text{Noise}\leq30\text{ mV}
\end{cases}
\end{equation}

\textbf{LDO:} The LDO test case features 27 design parameters: 10 transistor widths, 10 lengths, 4 ratios, 2 resistances, and 1 capacitance. With a design space of approximately $10^{69}$ sampling points, it is the largest among all test cases. The target specifications are:
\begin{equation}
C=
\begin{cases}
    \Delta\text{V}\leq0.1\text{ V,}&\text{Setup Time}\leq15\text{ ns}\\
    \text{PSRR}\geq60\text{ dB,}&\text{Noise}\leq5\text{ uV/}\sqrt{\text{Hz}}\\
    \text{Dropout}\leq 150 \text{ mV}
\end{cases}
\end{equation}

\textbf{SACMP:} The StrongArm latch comparator is a dynamic circuit with 14 design parameters: 6 transistor widths, 6 lengths, and 2 capacitances. It includes both symmetrical and cross-coupled structures. The target specifications are:
\begin{equation}
C=
\begin{cases}
    \text{Power}\leq40\text{ uW,}&\text{Set Delay}\leq4\text{ ns}\\
    \text{Reset Delay}\leq4\text{ns,}&\text{Noise}\leq120\text{ uV}
\end{cases}
\end{equation}

We apply TopoSizing to all four test cases and record both the number of simulation samples and the runtime. To provide a comprehensive performance comparison, we also evaluated commercial auto-sizing tools from Cadence Virtuoso\cite{Vir}, as well as two academic baselines: TuRBO \cite{Turbo} and a reinforcement learning–based approach\cite{PVTSizing}, under the same design specifications.

\subsection{Implementation Details}
We evaluate the proposed LLM-guided circuit optimization framework through a series of experiments encompassing baseline comparisons, recent LLM-aided approaches, and an ablation study. All optimization tasks are performed using TuRBO implemented in PyTorch, with circuit simulations conducted in Cadence Spectre under the SMIC 55\,nm PDK. Each circuit is optimized over ten independent runs to mitigate stochastic effects, and the average results are reported. The language model adopted in all LLM-related settings is GPT-4o\cite{GPT4} accessed via the OpenAI API, with the generation temperature fixed at 0.5. Unless otherwise specified, no fine-tuning or task-specific pretraining is applied.

For circuit understanding, we rely on graph-based processing. Circuit netlists are first converted into graph representations, where devices and terminals correspond to nodes and electrical connections to edges. We employ \texttt{networkx} to support sub-circuit matching via subgraph isomorphism, as well as path-finding algorithms for identifying connectivity patterns between modules. This enables systematic construction of the component--module--stage hierarchy. All intermediate graph structures are stored in a standardized JSON-like format, which preserves device attributes, connectivity, and hierarchy for subsequent reasoning and optimization.

The allocation of design parameters (e.g., widths, lengths, finger numbers) to each device is supposed to be done by LLMs. Within our framework, this assignment is automatically inferred during circuit understanding and achieves complete correctness across all benchmarks. In contrast, prior LLM-based approaches that directly process raw netlists often fail to achieve correct parameter assignment, leading to ambiguous or inconsistent sizing configurations. To ensure fairness in our comparisons, we manually complete the parameter assignment step before optimization when evaluating these baselines. This guarantees that all methods begin the sizing stage from an equally valid configuration, and that the performance differences reflect only the efficiency of the LLM intervention strategy.

For non-LLM baselines, we consider a pure TuRBO implementation without language model guidance, a reinforcement learning–based optimization strategy~\cite{PVTSizing}, and the built-in optimizer provided by Cadence Virtuoso. For LLM-based baselines, all methods employ GPT-4o in combination with TuRBO to ensure fairness, while differing only in the intervention strategy. Specifically, we compare our proposed adaptive intervention with a variant where the model intervenes at every optimization iteration~\cite{LEDRO}, as well as the ADO-LLM approach in which two candidate points are proposed at each round~\cite{ADOLLM}.

To better reflect practical design requirements, experiments are conducted under two distinct scenarios: constraint satisfaction optimization, which focuses on rapidly identifying feasible solutions, and single-objective optimization, which aims to achieve progressive performance improvement over multiple iterations. In the ablation study, we assess the contributions of two key components by selectively removing either the topology analysis module or the LLM-guided enhancement to the TuRBO process, while keeping all other settings fixed.

\subsection{Optimization Formulation}
\label{sec:formulation}
We now present the mathematical formulation of the optimization problems considered in our experiments. 
We first introduce the notation for the design parameters, performance metrics, performance specifications, and optimization directions:
\[
\begin{aligned}
\mathbf{X} &= (X_1,\dots,X_n) \in \mathcal{D} \subseteq \mathbb{R}^n, \\
\mathbf{F}(\mathbf{X}) &= (F_1(\mathbf{X}),\dots,F_m(\mathbf{X})), \\
\mathbf{C} &= (C_1,\dots,C_m), \\
\boldsymbol{\varphi} &= (\varphi_1,\dots,\varphi_m), \quad \varphi_i \in \{-1, 1\}.
\end{aligned}
\]
Here, $\mathbf{X}$ denotes the $n$-dimensional vector of design parameters, and $\mathcal{D}$ is the corresponding design space. $\mathbf{F}(\mathbf{X})$ is the $m$-dimensional vector of performance metrics obtained from circuit simulation, and $\mathbf{C}$ is the $m$-dimensional vector of performance specifications. The sign $\varphi_i$ encodes the optimization direction:
\[
\varphi_i=
\begin{cases}
    1, &\quad \text{for metrics to be maximized},\\
    -1, &\quad \text{for metrics to be minimized}.
\end{cases}
\]

\subsubsection{General normalized score}
For unified evaluation of metrics with different optimization directions, we define the normalized score of the $i$-th metric as:
\begin{equation}
r_i(\mathbf{X}) =\varphi_i\cdot
\frac{F_i(\mathbf{X}) - C_i}{\max(|F_i(\mathbf{X})|,\,|C_i|)},
\label{eq:ri}
\end{equation}
where the numerator adjusts the sign according to the optimization direction, and the denominator normalizes the scale. A positive $r_i$ means the $i$-th metric exceeds its specification, while a negative value indicates violation.

\subsection{Multi-constraint feasibility optimization}
For a pure feasibility problem, we focus on minimizing violations. The FoM is defined as:
\begin{equation}
\text{FoM}_{\text{feas}}(\mathbf{X}) =
\sum_{i=1}^m \min\big(0,\, r_i(\mathbf{X})\big),
\label{eq:fom_feas}
\end{equation}
so that each constraint contributes its violation magnitude if unsatisfied, and $0$ otherwise. The optimal feasible point achieves $\text{FoM}_{\text{feas}} = 0$.

\subsubsection{Single-objective constrained optimization}
When optimizing a specific target metric $F_t(\mathbf{X})$ while satisfying all constraints, we separate the non-target penalties from the target reward. We first quantify the penalties from all non-target metrics:
\begin{equation}
R_{\text{non-target}}(\mathbf{X}) = 
\sum_{\substack{i=1 \\ i\ne t}}^m \min\big(0,\, r_i(\mathbf{X})\big),
\label{eq:penalty_non_target}
\end{equation}
where each non-target metric contributes its violation magnitude if unsatisfied, and $0$ otherwise.

The contribution of the target metric $F_t(\mathbf{X})$ depends on whether all non-target constraints are satisfied:
\begin{equation}
R_{\text{target}}(\mathbf{X}) =
\begin{cases}
r_t(\mathbf{X}), & \text{if } r_i(\mathbf{X}) \ge 0 \ \forall i\ne t, \\
\min\big(0,\, r_t(\mathbf{X})\big), & \text{otherwise}.
\end{cases}
\label{eq:reward_target}
\end{equation}
This ensures that the target metric receives a positive reward only when all other constraints are satisfied; otherwise it is also penalized if it fails to meet its own specification.

Finally, the overall figure of merit is obtained by combining the two terms:
\begin{equation}
\text{FoM}_{\text{single}}(\mathbf{X}) =
R_{\text{non-target}}(\mathbf{X}) + R_{\text{target}}(\mathbf{X}).
\label{eq:fom_single}
\end{equation}

\subsection{Constraints Satisfaction}
Tab.~\ref{tab:final_results} summarizes the optimization results, averaged over ten independent runs. Compared with traditional sizing methods, our framework achieves clear improvements in both sampling efficiency and convergence speed across most test cases. Compared to other LLM-aided approaches, TopoSizing attains higher optimization efficiency while requiring substantially fewer LLM calls. Although our framework introduces additional preprocessing and circuit-understanding steps, the overall runtime remains lower, making this trade-off both practical and acceptable. 

\begin{figure*}[h]
  \raggedright 
  \includegraphics[width=1\textwidth]{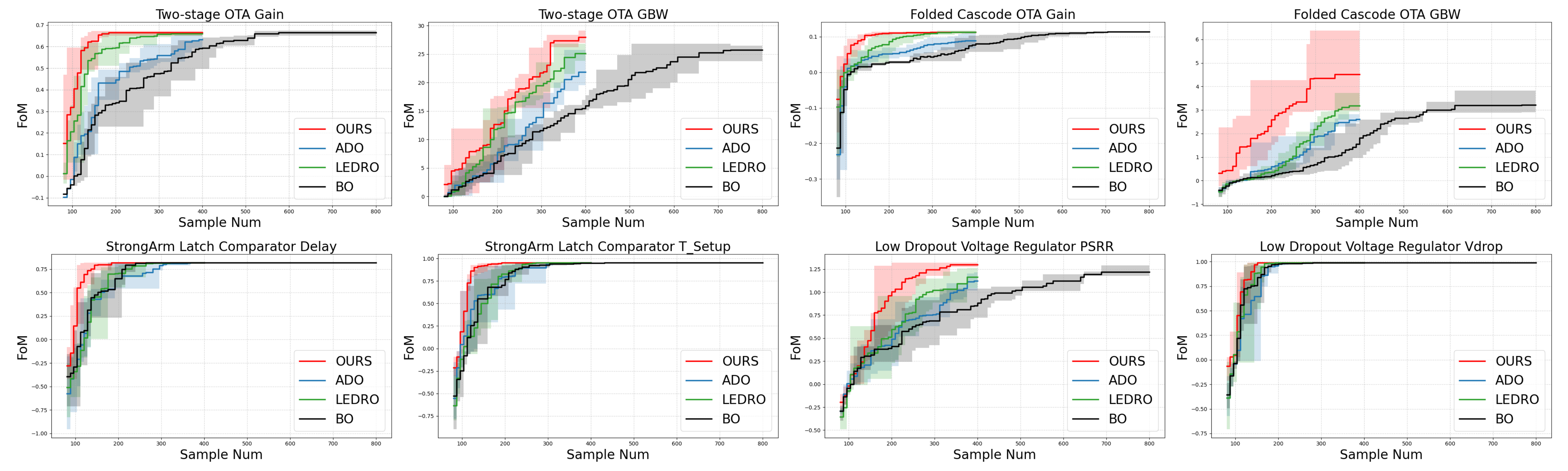}
  \caption{FoM versus sample count in the single-objective optimization, with curves representing different methods and shaded areas showing variation across runs.}
  \label{fig:single_object}
\end{figure*}
\subsection{Single-object Optimization}
We further evaluate each test case by selecting a single performance metric for single-objective optimization, in order to examine whether the proposed framework can not only accelerate optimization but also establish a correct mapping from design parameters to performance metrics. This experiment is designed to test whether LLMs, when guided through our framework, can capture such relationships more reliably and thereby enhance performance in larger-scale optimization tasks. As baselines, we compare against Bayesian Optimization (BO) with 400 samples, an LLM-based method with 400 samples, and BO with 800 samples. The experimental results confirm our expectation: except for the relatively simple 2stage case—where the LLM’s interpretation is already sufficiently accurate and thus improvements are less pronounced—our framework consistently achieves comparable or better optimization outcomes with fewer simulations.  

Closer inspection of the curves reveals three important observations. (1) Our framework starts from a higher FoM immediately after the initial sampling stage, indicating that the LLM-guided initialization provides a strong advantage. (2) Long stagnation phases rarely occur, as our intervention strategy effectively invokes the LLM to adjust the search space once stagnation is detected. (3) Comparing the baselines ADO and LEDRO further illustrates the role of understanding quality: in cases where the LLM’s interpretation is reliable (e.g., 2stage and FC), region-based pruning of the design space outperforms point-wise proposal strategies; in contrast, when the interpretation is less reliable (e.g., LDO), region-based modification may amplify errors and lead to performance degradation. These observations together validate both the effectiveness and the limitations of LLM-guided optimization, while highlighting the advantages of our proposed integration strategy.  

\subsection{Ablation Study}
Tab.~\ref{tab:ablation_1} presents the results of the ablation study, which demonstrate the effectiveness of the proposed TopoSizing framework. The analysis shows that both the topological information extraction and the iterative understanding by LLM contribute to improving circuit understanding accuracy. Furthermore, the enhanced understanding translates into higher optimization efficiency, thereby validating the overall effectiveness of our framework.

\begin{table}[t]
\centering
\caption{Optimization results on four real-world circuits: Comparison With Conventional Methods}
\label{tab:final_results}
\begin{threeparttable} 
\small
\setlength{\tabcolsep}{4pt}
\begin{tabular}{l l *{4}{S[table-format=4.0]}}
\toprule
\multicolumn{1}{c}{\multirow{2}{*}{Category}} & 
\multicolumn{1}{c}{\multirow{2}{*}{Algorithm}} & 
\multicolumn{4}{c}{Test Cases} \\
\cmidrule(l){3-6}
& & {OTA} & {FCOTA} & {SACMP} & {LDO} \\
\midrule

\multirow{4}{*}{Sample \#} 
& Proposed      & \textbf{18}  & \textbf{32}  & \textbf{18}  & \textbf{22}  \\
& TuRBO      & {25}  & {66}  & {107}  & {37}  \\
& RL   & {49} & {86} & {86} & {60} \\
& Virtuoso    & {197} & {142}  & {108} & {400}  \\
& \textbf{Eff. impr.} & \textbf{ 1.4$\times$} & \textbf{ 2.2$\times$} & \textbf{ 4.8$\times$} & \textbf{ 1.4$\times$} \\
\midrule

\multirow{4}{*}{Runtime (s)} 
& Proposed      & \textbf{167} & \textbf{608} & \textbf{273} & \textbf{234} \\
& TuRBO      & {234} & {1270} & {1907} & {412} \\
& RL   & {2178} & {4705} & {4407} & {6553} \\
& Virtuoso  & {324} & {947} & {950} & {1141} \\
& \textbf{Speed up} & \textbf{1.2$\times$} & \textbf{1.5$\times$} & \textbf{3.5$\times$} & \textbf{1.4$\times$} \\

\bottomrule
\end{tabular}

\end{threeparttable}
\end{table}

\begin{table}[t]
\centering
\caption{Results of Ablation Study}
\label{tab:ablation_1}
\begin{threeparttable}

\small
\setlength{\tabcolsep}{4pt}
\begin{tabular}{l l *{4}{c}}
\toprule
\multicolumn{1}{c}{\multirow{2}{*}{Category}} & 
\multicolumn{1}{c}{\multirow{2}{*}{Algorithm}} & 
\multicolumn{4}{c}{Test Cases} \\
\cmidrule(l){3-6}
& & {OTA} & {FCOTA} & {SACMP} & {LDO} \\
\midrule
\multirow{3}{*}{Sample \#} 
& Proposed      & \textbf{18}  & \textbf{32}  & \textbf{18}  & \textbf{22}  \\
& W/O IU$^*$   & {19} & {52} & {69} & {39} \\
& W/O TIE$^\dagger$   & {37} & {66} & {107} & {37} \\
\midrule

\multirow{3}{*}{\shortstack{Accuracy of \\ Classification}} 
& Proposed & \textbf{100\%} & \textbf{100\%} & \textbf{100\%} & \textbf{100\%} \\
& W/O IU   & {100\%} & {87\%} & {89\%} & {86\%} \\
& W/O TIE    & {100\%} & {48\%} & {58\%} & {59\%} \\

\bottomrule
\end{tabular}

\begin{tablenotes} 
\footnotesize
    \item[*] Iterative understanding by LLM
    \item[$\dagger$] Topological Information Extraction
\end{tablenotes}
\end{threeparttable}
\end{table}

\begin{table}[t]
\centering
\caption{Optimization results on four real-world circuits: Comparison With LLM-aided Methods}
\label{tab:final_results}
\begin{threeparttable} 
\small
\setlength{\tabcolsep}{4pt}
\begin{tabular}{l l *{4}{S[table-format=4.0]}}
\toprule
\multicolumn{1}{c}{\multirow{2}{*}{Category}} & 
\multicolumn{1}{c}{\multirow{2}{*}{Algorithm}} & 
\multicolumn{4}{c}{Test Cases} \\
\cmidrule(l){3-6}
& & {OTA} & {FCOTA} & {SACMP} & {LDO} \\
\midrule

\multirow{4}{*}{Sample \#} 
& Proposed      & {18}  & \textbf{32}  & \textbf{18}  & \textbf{22}  \\
& LEDRO      & {18}  & {59}  & {60}  & {33}  \\
& ADO-LLM   & {24} & {55} & {72} & {35} \\
& \textbf{Eff. impr.} & \textbf{ 1.0$\times$} & \textbf{ 1.6$\times$} & \textbf{ 2.8$\times$} & \textbf{ 1.5$\times$} \\
\midrule

\multirow{4}{*}{Runtime (s)} 
& Proposed      & \textbf{167} & \textbf{608} & \textbf{273} & \textbf{234} \\
& LEDRO      & {172} & {1328} & {836} & {407} \\
& ADO-LLM   & {233} & {1081} & {972} & {427} \\
& \textbf{Speed up} & \textbf{1.0$\times$} & \textbf{1.8$\times$} & \textbf{3.6$\times$} & \textbf{1.7$\times$} \\
\midrule

\multirow{4}{*}{LLM Calls \# } 
& Proposed & \textbf{1.1} & \textbf{3.3} & \textbf{2.2} & \textbf{2.1} \\
& LEDRO    & {2.3} & {7.4} & {7.5} & {4.1} \\
& ADO-LLM   & {3.0} & {7.0} & {9.0} & {4.4} \\
\bottomrule
\end{tabular}
\end{threeparttable}
\end{table}

\subsection{Case Study}
\begin{figure}[t]
  \raggedright 
  \includegraphics[width=0.48\textwidth]{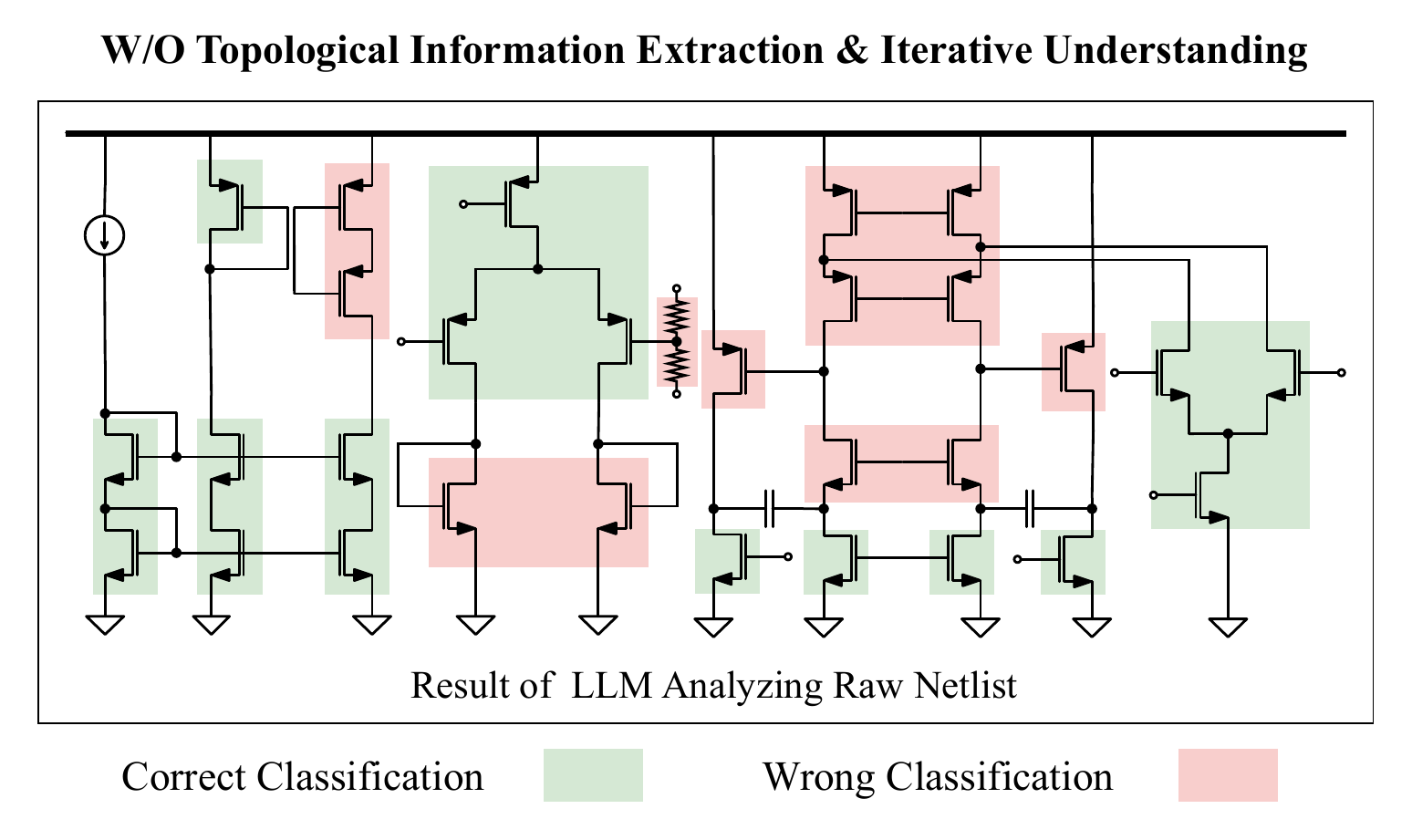}
  \caption{LLM processing raw netlist }

  \label{fig:case_study_llm}
\end{figure}
\begin{figure*}[]
\includegraphics[width=0.98\textwidth]{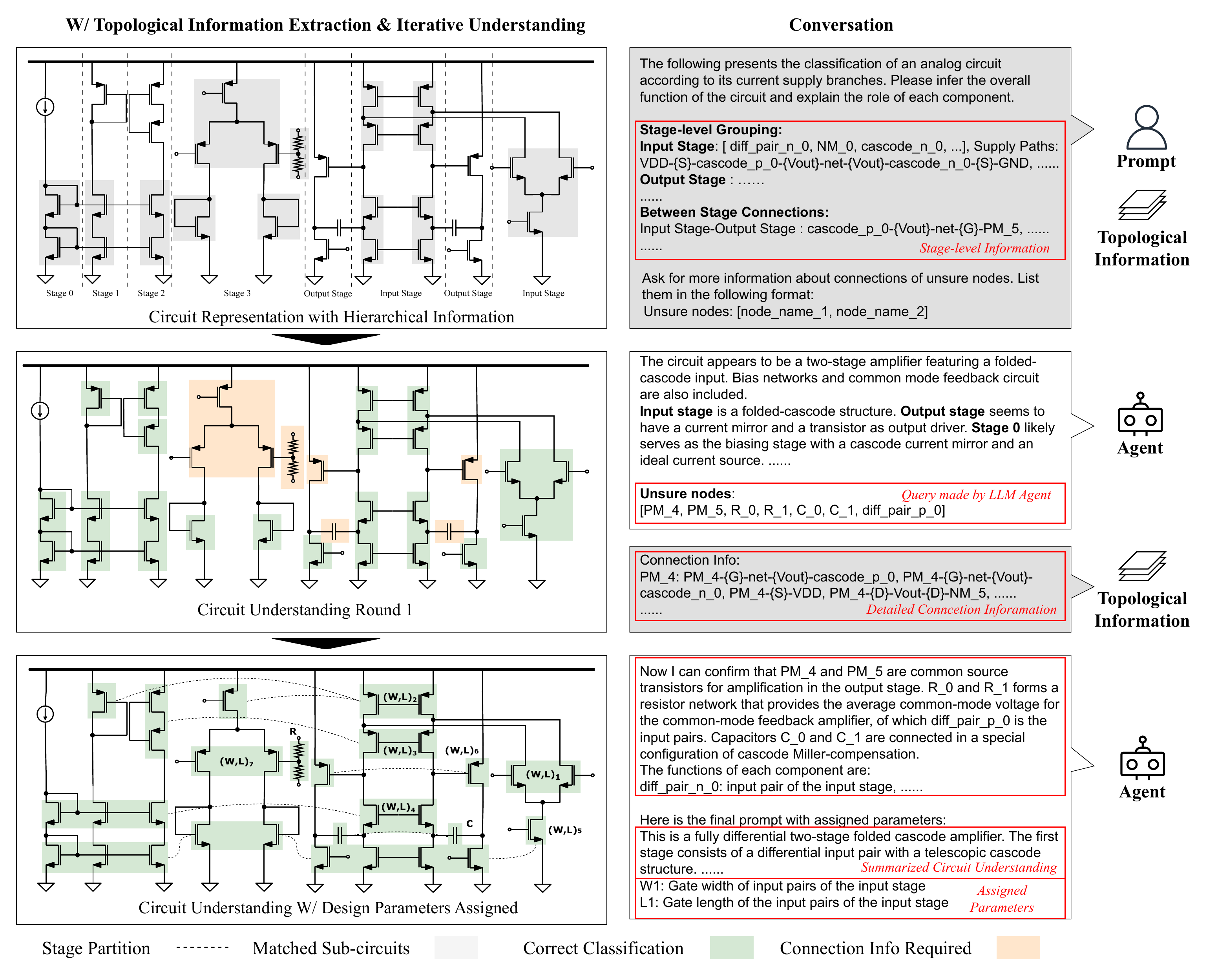}
  \caption{ The detailed steps of our iterative circuit understanding process and final output with parameter assigned. }

  \label{fig:case_study}
\end{figure*}
Finally, we conduct a case study on the most complex FCOTA circuit to evaluate the reliability of circuit understanding in a concrete setting. As shown in Fig.~\ref{fig:case_study_llm}, when directly reading the raw netlist, the LLM produces incorrect functional interpretations, failing to capture key structural relationships. In particular, for the FCOTA with a common-mode feedback (CMFB) module, the model—without any auxiliary structure—misidentifies devices around the output node(s), conflating CMFB-related elements with the main signal path. This leads to an erroneous assignment of roles in the first stage (e.g., mislabeling the load/sleeve devices and the input pair), which propagates downstream. In contrast, after applying our framework (Fig.~\ref{fig:case_study}), the hierarchical preprocessing and iterative verification (topological extraction, conduction-graph filtering, port-role inference, and grouping) enable the model to disentangle the CMFB loop from the differential signal path and to recover a complete and correct stage-level interpretation. This comparison highlights that our framework not only improves accuracy but also ensures robustness, eventually reaching 100\% correctness through repeated verification.

\section{Conclusion}
We propose TopoSizing, a stable and efficient topology-aware framework for analog circuit sizing. TopoSizing accelerates the optimization process through LLM integration for circuit topology comprehension and design space pruning. Experiments across four real-world test cases demonstrate its superior sampling efficiency and time efficiency compared to prior sizing tools from both industry and academia.

This fully automated and efficient design framework demonstrates significant practical value, offering a novel approach to accelerate traditional analog sizing processes through the integration of advanced AI capabilities.

\clearpage


\end{document}